\documentclass{article}

\usepackage[preprint]{neurips_2026}

\usepackage{enumitem}
\usepackage[utf8]{inputenc} 
\usepackage[T1]{fontenc}    
\usepackage{hyperref}       
\usepackage{url}            
\usepackage{booktabs}       
\usepackage{amsfonts}       
\usepackage{nicefrac}       
\usepackage{microtype}      
\usepackage{xcolor}         
\usepackage{graphicx}
\usepackage{multirow}
\usepackage{caption}
\usepackage{wrapfig}
\usepackage{subcaption}
\usepackage{enumitem}
\usepackage{wrapfig}
\usepackage{etoc}
\etocdepthtag.toc{mtchapter}
\etocsettagdepth{mtchapter}{subsection}
\etocsettagdepth{mtappendix}{none}

\usepackage{algorithmic}
\usepackage[ruled,vlined,linesnumbered]{algorithm2e}
\usepackage{etoolbox}

\usepackage{hyperref}
\hypersetup{
    colorlinks=true,
    linkcolor=blue,
}
\usepackage{xcolor}
\usepackage{color, soul}

\usepackage[normalem]{ulem}
\useunder{\uline}{\ul}{}
\title{GRAFT: Graph-Tokenized LLMs for Tool Planning}

\author{
  Xinyi Gao$^1$, Xinyu Ren$^1$, Junliang Yu$^2$, Tong Chen$^1$, Quoc Viet Hung Nguyen$^2$, Hongzhi Yin$^1$ \\ \\ 
  $^1$The University of Queensland, $^2$ Griffith University
}

\begin{document}

\maketitle

\begin{abstract}
Large language models (LLMs) are increasingly used to complete complex tasks by selecting and coordinating external tools across multiple steps. This requires aligning tool choices with subtask intent while satisfying directional execution dependencies among tools. To do this, existing methods model these dependencies as tool graphs and incorporate the graphs with LLMs through retrieval, serialization, or prompt-level injection. However, these external graph-use strategies all follow a matching paradigm, which often fails to align tool choices with the underlying subtask structure, producing semantically plausible plans that violate graph constraints. This issue is further exacerbated by error accumulation, where an early incorrect tool selection shifts the plan into an invalid graph state and causes subsequent predictions to drift away from the valid execution path. To address these challenges, we propose \textbf{GRAFT}, a graph-tokenized language model framework for dependency-aware tool planning. GRAFT internalizes the tool graph by mapping each tool node to a dedicated special token and learning directed tool dependencies within the representation space. It further introduces on-policy tool context distillation, training the model on its own sampled trajectories while distilling stepwise planning signals. Experiments show that GRAFT achieves state-of-the-art performance in exact sequence matching and dependency legality, supporting more reliable LLM tool planning in complex workflows.

\end{abstract}

\section{Introduction}

Large language models (LLMs) are increasingly extending language understanding into tool-augmented problem solving, where they interact with external tools, APIs, and expert models to perform precise computation, and support domain-specific operations beyond language generation alone~\cite{shen2024llm, xu2025llm}.
Despite the growing role of LLMs in task solving, multi-step tool planning remains a central challenge. Given a high-level user instruction, the model must understand the task intent, decompose it into subtasks, and generate an executable tool sequence in a valid order~\cite{wu2024can}. This process is inherently dependency-aware, since tools are rarely independent and the output of one tool may serve as the input, prerequisite, or semantic context for another. Such dependencies naturally define a \textit{tool dependency graph}~\cite{liu2024toolnet}, where nodes denote tools and directed edges specify valid execution transitions. Therefore, reliable multi-step tool planning requires generating tool sequences that satisfy both task semantics and graph-structured execution constraints, making it a dependency-aware sequence generation problem over a directed tool graph.

\begin{figure*}[t]
\centering
\includegraphics[width=\linewidth]{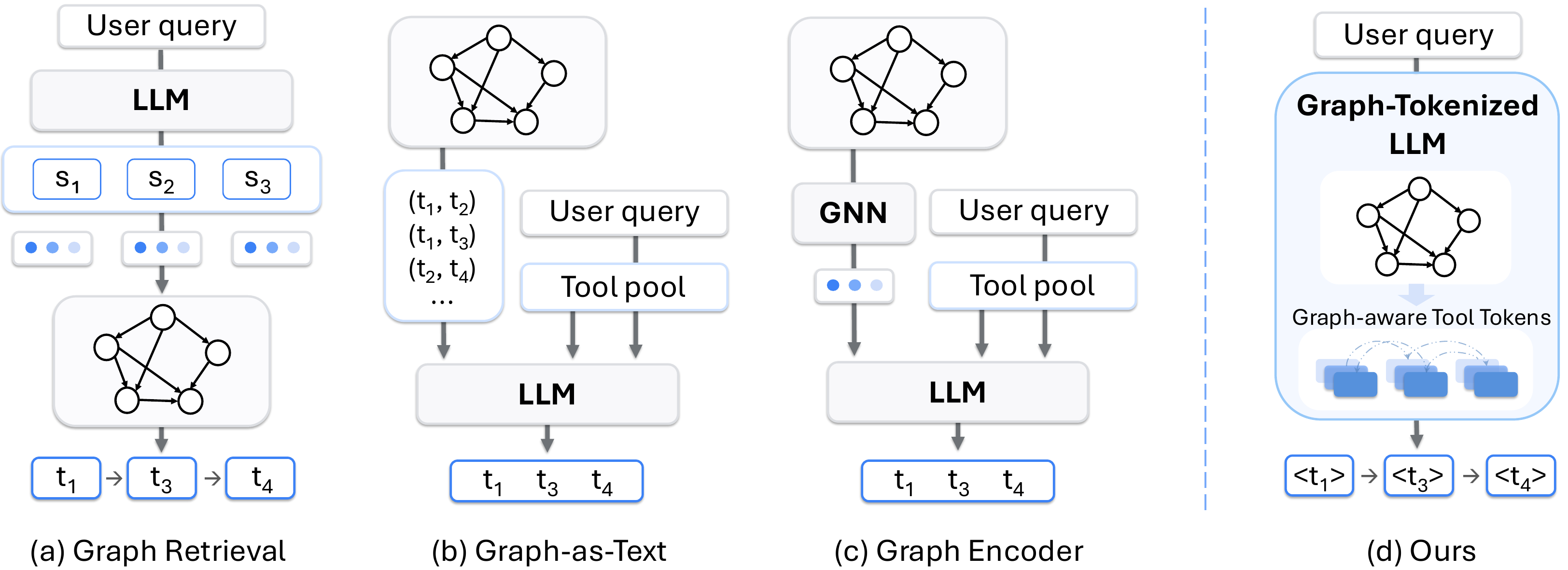}
\caption{Existing methods use the tool graph externally: they retrieve tools $t$ for decomposed subtasks $s$, serialize the graph into text, or inject graph embeddings as auxiliary LLM inputs. In contrast, our method parameterizes the tool graph as dependency-aware tool tokens, enabling the LLM to directly generate tool plans from the user query.}
\label{mainfig}
\end{figure*}

While this formulation highlights the importance of tool dependencies, existing methods still rely on external mechanisms to connect task reasoning with tool selection. A common strategy~\cite{10.1007/978-3-031-73254-6_6,wu2024can} is to first prompt the LLM to decompose the user instruction into intermediate subtasks and then retrieve tools for each subtask through graph search, as shown in Fig. \ref{mainfig} (a). However, this separates task decomposition from tool selection, so the generated subtasks may be incomplete or mismatched with available tools to construct a valid execution path. Another line of work, illustrated in Fig. \ref{mainfig} (b-c), provides graph structure as auxiliary context, either by serializing the tool graph into text~\cite{pmlr-v235-lin24k, wu2024can} or by encoding tool relations into dense representations~\cite{chen2026gtool}. 
However, both types of methods require lengthy prompts and struggle to align subtask relationships with tool transition semantics.
Moreover, the core of all existing methods are still ``matching'', which heavily depends on the quality of query parsing and graph representation. 
Consequently, such external graph context may inform the LLM but cannot reliably constrain each next-tool decision, often producing semantically plausible plans that violate structural dependencies~\cite{wu2024can}.

Beyond these structural limitations, LLM-based tool planning is also vulnerable to error accumulation. During multi-step planning, an early incorrect tool prediction changes the semantic state of the plan and moves generation away from the intended execution path. Subsequent predictions are then conditioned on an imperfect prefix, causing errors to propagate across later steps. This issue is exacerbated by supervised fine-tuning (SFT), which optimizes the model under gold prefixes during training but requires it to condition on its own generated prefixes during inference. Such train-test mismatch, known as exposure bias~\citep{bengio2015scheduled,ranzato2016sequence,lamb2016professor}, reduces robustness to off-path states and leads to cascading failures in multi-step tool planning.

In light of these challenges, we propose Graph-Tokenized LLMs for Tool Planning (\textbf{GRAFT}), a framework that internalizes directed tool dependencies into LLMs and improves planning robustness through subtask-aware on-policy distillation. 
We first map each tool node to a dedicated special token and train the model to reconstruct directed successor relations from the hidden states of tool tokens. 
This \textbf{Graph-Token Edge Reconstruction} objective turns the tool graph from an external retrieval target into a parameterized capability of the LLM, enabling the model to jointly learn tool semantics and executable dependency structure. 
To mitigate exposure bias, we introduce \textbf{On-Policy Tool Context Distillation}.
For each sample, the student samples a tool-token trajectory from its current policy using only the user query.
Conditioned on the student-generated prefix, a frozen teacher is initialized from the same model and receives decomposed subtasks as privileged context to provide next-tool supervisions.
This yields a self-distillation objective~\cite{zhao2026self, ye2026policy} that trains the student on its own on-policy states while internalizing subtask-context guidance, without requiring such context at inference time.
As a result, GRAFT learns to predict tools according to both the directed dependency graph and the intended reasoning process, leading to more executable and efficient multi-step planning.

Our contributions are summarized as follows:
\begin{itemize}

    \item We identify the limitations of external graph usage and SFT under exposure bias in multi-step tool planning, and propose a new formulation that internalizes the directed tool graph into LLMs for efficient dependency-aware tool planning.
    
    \item We propose a graph-token learning framework that internalizes directed tool dependencies into LLMs through tool-specific tokens and graph-token edge reconstruction, enabling the model to learn both tool semantics and executable transitions.
    
    \item We introduce on-policy tool context distillation, which uses subtasks as teacher planning context and trains the model on its own sampled trajectories to learn task decomposition signals for robust tool sequence generation.
    
    \item Experiments show that GRAFT achieves state-of-the-art performance, outperforming retrieval-, serialization- and prompt-based baselines in exact sequence matching and dependency legality, advancing the development of more reliable LLMs for practical workflows.
\end{itemize}

\section{Related Work}
\subsection{LLM-based Tool Planning}

Early LLM-based methods mainly formulate tool planning as subtask-tool matching. They decompose a user query into subtasks and select suitable tools through prompting~\cite{shen2023hugginggpt,shen2024taskbench,qin2023toolllm}, retrieval~\cite{qin2023toolllm,yang2023gpt4tools}, or tool-token generation~\cite{wang2025toolgen}. 
More recently, agentic tool-use methods~\cite{masterman2024landscape} formulate planning as a closed-loop decision process, where the LLM selects the next tool action based on the previous context, and execution observations, e.g., ReAct~\cite{yao2022react}, AgentGym~\cite{xi2025agentgym}, and AgentFlow~\cite{li2025flow}. 
This interaction-based paradigm can refine decisions after tool execution, but it relies on a strong backbone LLM and requires repeated LLM calls with observation updates. More importantly, all these methods do not explicitly model directed dependencies among tools, making it difficult to guarantee dependency-valid tool sequences before execution.

To capture tool dependencies, recent graph-based studies represent tools as nodes and dependency relations as directed edges. ControlLLM~\cite{10.1007/978-3-031-73254-6_6} and PLAG~\cite{pmlr-v235-lin24k,wu2024can} search solution paths over the tool graph, GNN4Plan~\cite{wu2024can} studies graph learning for task planning, and GTool~\cite{chen2026gtool} injects query-specific graph representations into LLMs. These methods demonstrate the importance of graph structure for valid tool sequencing. However, the graph is still used as an external search space, reasoning module, or auxiliary context, rather than being internalized into the LLM generation process.

In contrast, this paper focuses on the fundamental capability of single-pass pre-execution planning. The LLM is required to understand the task and generate a complete dependency-valid tool sequence before any tool is executed, without relying on post-call observations or repeated interaction. This provides a reliable and efficient planning foundation for downstream LLM-based agents.

\subsection{On-policy Optimization}

To adapt LLMs to task-specific generation, supervised fine-tuning (SFT) is commonly used to train the model on gold prefixes. However, during inference, the model must condition on its own generated prefixes, leading to a train-test mismatch known as exposure bias \cite{bengio2015scheduled,ranzato2016sequence,lamb2016professor}. 
On-policy learning mitigates this mismatch by optimizing the model on trajectories sampled from its current policy, thereby exposing it to inference-time states. 
This idea has been explored in Scheduled Sampling~\citep{bengio2015scheduled},
DAgger~\citep{ross2011dagger}, on-policy reinforcement learning methods ( e.g.,
REINFORCE~\citep{williams1992reinforce}, TRPO~\citep{schulman2015trpo},
and PPO~\citep{schulman2017ppo}) and on-policy distillation methods~\citep{song2026survey, ye2026policy, zhao2026self}, all of which aim to improve robustness under policy-induced state distributions.
In multi-step tool planning, exposure bias is particularly harmful, where an early incorrect tool prediction can move the model away from the intended execution path and cause subsequent errors to accumulate. 
Therefore, on-policy learning provides an important paradigm for improving the robustness and dependency validity of generated tool sequences.

\section{Methodology}

To address the aforementioned challenges, we propose a graph-tokenized framework for executable multi-step tool planning. 
Our framework first internalizes directed tool dependencies into the LLM by representing tools as dedicated tokens and learning graph-aware tool representations. 
It then improves multi-step planning robustness through on-policy distillation, enabling the model to generate complete dependency-valid tool sequences under its own autoregressive planning states.

\subsection{Problem Formulation}
Let $\mathcal{T}=\{t_1,t_2,\ldots,t_M\}$ denote the tool set, where each tool $t_i$ is an executable API or function with textual metadata, such as its name and description.
The dependencies among tools are represented as a directed tool graph $\mathcal{G}=(\mathcal{T},\mathcal{E})$, where each directed edge $(t_i,t_j)\in\mathcal{E}$ indicates that tool $t_j$ can be legally invoked after $t_i$.
Such a dependency typically means that the output or precondition produced by $t_i$ is required by $t_j$.

Given a user query $q$, the goal of tool planning is to generate an executable tool trajectory that completes the task by invoking tools in a valid order, as illustrated in Fig.~\ref{dataset_examples}.
A training sample is denoted as $(q,s,y)\in\mathcal{D}$, where $s=(s_1,\ldots,s_L)$ denotes the pre-defined subtask decomposition provided in the training data, and $y=(y_1,\ldots,y_L)$ is the corresponding ground-truth tool trajectory.
Each element $y_k\in\mathcal{T}$ denotes the tool used to solve subtask $s_k$.
The ground-truth trajectory is executable under the directed tool graph, satisfying
\begin{equation}
(y_k,y_{k+1})\in\mathcal{E}, \quad k=1,\ldots,L-1.
\label{eq:gold_trajectory}
\end{equation}
We study LLM-based tool planning, where an LLM $\pi_{\theta}$ takes the user query $q$ and the directed tool graph $\mathcal{G}$ as input, and generates a predicted tool trajectory $\hat{y}=(\hat{y}_1,\ldots,\hat{y}_K)$ as: 
\begin{equation}
\pi_{\theta}(\hat{y}\mid q,\mathcal{G})
=
\prod_{k=1}^{K}
\pi_{\theta}(\hat{y}_k \mid q,\mathcal{G},\hat{y}_{<k}).
\label{eq:tool_policy}
\end{equation}
The expected output is a tool sequence that matches the ground-truth trajectory $y$ and satisfies the tool dependencies specified by $\mathcal{G}$.

\subsection{Graph Tokenization and Subtask Grounding}

To internalize the tool graph into the LLM rather than using it as external context, we first treats each tool as an atomic semantic unit by representing each tool node as a dedicated token.

\paragraph{Tool tokenization.}
We map each tool $t_i \in \mathcal{T}$ to a dedicated token via an indexing function $\tau$:
\begin{equation}
\tau(t_i)=a_i,
\quad
\mathcal{A}=\{a_1,a_2,\ldots,a_M\},
\quad
\mathcal{V}'=\mathcal{V}_{\mathrm{LM}}\cup\mathcal{A}.
\label{eq:tool_token_vocab}
\end{equation}
Here, $\mathcal{V}_{\mathrm{LM}}$ denotes the original LLM vocabulary, $\mathcal{A}$ is the tool-token set, and $\mathcal{V}'$ is the expanded vocabulary.
We adopt an atomic token format, where each tool is represented as a single indivisible token, e.g., the \texttt{Image-to-Text} tool is mapped to \texttt{<Image\_to\_Text>}. The embedding of each tool token is initialized using its textual metadata.

Afterwards, for a ground-truth tool trajectory $y=(y_1,\ldots,y_L)$, we can tokenize it as token sequence $z=(z_1,\ldots,z_L)$, where $z_k=\tau(y_k)\in\mathcal{A}$.
Tool planning is then converted into autoregressive generation over $\mathcal{A}$.
This formulation constrains generation to the tool-token space rather than free-form text, thereby reducing hallucinated tool invocations as our empirical study in Appendix~\ref{app_hall}.

\paragraph{Subtask-to-tool grounding.}
Tool metadata provides only static descriptions, whereas the same tool may serve different purposes across planning scenarios.
To help the LLM associate tool tokens with their practical usage in specific task contexts, we construct subtask-token pairs from each training example $(q,s,y)$:
\begin{equation}
\mathcal{D}_{\mathrm{sub}}
=
\{(s_k,z_k) \mid (q,s,y)\in\mathcal{D},\ z_k=\tau(y_k),\ k=1,\ldots,|y|\}.
\label{eq:subtask_pairs}
\end{equation}
Here, the subtask $s_k$ describes the local operation required at step $k$, while the tool token $z_k$ represents the tool selected to perform it.
The model is then trained to generate the corresponding tool token conditioned on each subtask:
\begin{equation}
\mathcal{L}_{\mathrm{sub}}
=
-
\sum_{(s_k,z_k)\in\mathcal{D}_{\mathrm{sub}}}
\log \pi_{\theta}(z_k\mid s_k).
\label{eq:subtask_loss}
\end{equation}
This grounding stage connects natural-language subtasks with executable tool tokens, enabling the model to learn task-specific tool usage. The resulting grounded tool-token representations then provide the basis for directed edge reconstruction in the next stage.

\subsection{Graph-Token Edge Reconstruction}

\paragraph{Training sequence generation.}
With the constructed tool tokens, we expect them to capture dependency relations in the tool graph.
A direct approach is to train each tool token independently to reconstruct its neighboring tools.
However, this requires separate forward passes for individual tool nodes, which is inefficient and inconsistent with autoregressive tool planning, where tools are generated conditioned on preceding tool contexts.
To align with this generation process, we construct training sequences $\rho$ by sampling directed paths with different lengths from the tool graph $\mathcal{G}$:
\begin{equation}
\rho=(u_1,\ldots,u_R),
\quad
u_r\in\mathcal{T},
\quad
(u_r,u_{r+1})\in\mathcal{E},
\label{eq:graph_walk}
\end{equation}
where $u_r$ denotes a tool node and $1\leq R\leq R_{\max}$. These variable-length paths provide sequential dependency contexts, ranging from single-tool states to multi-hop dependencies.

\paragraph{Directed edge reconstruction.}
Given a sampled path $\rho$, we feed the corresponding tool-token sequence into the LLM. 
For each position $r$, we extract the contextual hidden state $\mathbf{h}_r$ from the penultimate layer at the tool-token position of $u_r$. 
We then construct a candidate set $\mathcal{C}(u_r)$ for $u_r$ containing its valid outgoing neighbors $\mathcal{N}^{+}(u_r)$, sampled negative tools, and reverse-edge negatives. 

For each tool $t_j\in\mathcal{C}(u_r)$, let $\mathbf{e}_j$ be the embedding of its tool token and the directional edge score is computed as
\begin{equation}
s_{rj}
=
\frac{
\mathrm{sim}
\left(
\mathbf{W}_{h}\mathbf{h}_{r},
\mathbf{W}_{e}\mathbf{e}_{j}
\right)
}{\gamma},
\label{eq:successor_score}
\end{equation}
where $\mathbf{W}_{h}$ and $\mathbf{W}_{e}$ are projection matrices, $\mathrm{sim}(\cdot,\cdot)$ is normalized dot product, and $\gamma$ is a temperature coefficient.
The reconstruction loss for the tools in path $\rho$ is defined as
\begin{equation}
\mathcal{L}_{\mathrm{edge}}(\rho)
=
-
\frac{1}{|\mathcal{I}_{\rho}|}
\sum_{r\in\mathcal{I}_{\rho}}
\frac{1}{|\mathcal{N}^{+}(u_r)|}
\sum_{t_j\in\mathcal{N}^{+}(u_r)}
\log
\frac{\exp(s_{rj})}
{\sum_{t_k\in\mathcal{C}(u_r)}\exp(s_{rk})},
\label{eq:edge_loss}
\end{equation}
where $\mathcal{I}_{\rho}=\{r\mid \mathcal{N}^{+}(u_r)\neq\emptyset\}$ denotes the positions whose tools have outgoing edges.
The final objective averages Eq. (\ref{eq:edge_loss}) over all sampled paths.

Instead of predicting only the next tool in the sampled path $\rho$, each node representation is trained to recover its graph-defined successors. 
This contrastive reconstruction objective encourages $\mathbf{h}_r$ to encode the outgoing dependency structure of $u_r$, allowing directed edges to be recovered from tool-token representations~\cite{li2023survey,zhang2026toward,jin2024large}. 
Thus, the LLM internalizes directed tool dependencies and leverages them at inference time without explicit graph serialization or retrieval.

\subsection{On-Policy Tool Context Distillation}

After internalizing the tool graph, the model captures both the semantic meanings of tool tokens and the directed dependency relations among tools. Then, we adapt it to query-conditioned tool planning through a two-stage training procedure.

\paragraph{Query-to-tool SFT.}
We first warm up the graph-tokenized policy with SFT, training it to generate the tool-token sequence from the user query.
For each training sample $(q,s,y)\in\mathcal{D}$ and the tokenized trajectory $z=(z_1,\ldots,z_L)$, the SFT objective is
\begin{equation}
\mathcal{L}_{\mathrm{sft}}
=
-
\sum_{(q,s,y)\in\mathcal{D}}
\sum_{k=1}^{L}
\log
\pi_{\theta}(z_k \mid q,z_{<k}).
\label{eq:sft_loss}
\end{equation}
This warm-up stage provides a stable query-to-tool planning policy for subsequent on-policy rollout.

\paragraph{On-policy tool context distillation.}

After SFT, we initialize both the student and teacher from the SFT checkpoint. The student is the trainable graph-tokenized policy $\pi_{\theta}$, while the teacher is a frozen copy of the same model, denoted by $\pi_{\bar{\theta}}$. Unlike the student, the teacher is additionally conditioned on privileged planning context, including decomposed subtasks and the ground-truth tool trajectory. This forms a self-distillation setting~\cite{song2026survey,zhao2026self} in which the teacher provides context-aware planning guidance, while the student learns to reproduce such guidance under its own generated tool prefixes.

Specifically, we define the student and teacher prompts as
\begin{equation}
P_S(q)
=
[\mathrm{Query}: q],
\quad
P_T(q,s,y)
=
[\mathrm{Query}: q;\ \mathrm{Subtasks}: s;\ \mathrm{Tools}: z].
\label{eq:distill_prompts}
\end{equation}
The student receives only $P_S(q)$ and samples a tool-token trajectory $\hat{z}\sim\pi_{\theta}(\cdot\mid P_S(q))$.
At step $k$, both policies are conditioned on the same student-generated prefix $\hat{z}_{<k}$ and produce next-token distributions:
\begin{equation}
p^S_k
=
\pi_{\theta}(\cdot\mid P_S(q),\hat{z}_{<k}),
\quad
p^T_k
=
\pi_{\bar{\theta}}(\cdot\mid P_T(q,s,y),\hat{z}_{<k}).
\label{eq:teacher_student_policy}
\end{equation}
Here, the subtasks provide step-level planning guidance, while the ground-truth trajectory anchors the teacher to the correct tool sequence, enabling token-level supervision under student-generated prefixes.
Finally, we minimize the reverse KL divergence~\cite{gu2024minillm,ye2026policy,song2026survey} along the student rollout:
\begin{equation}
\mathcal{L}_{\mathrm{opd}}
=
\mathbb{E}_{(q,s,y)\sim\mathcal{D}}
\mathbb{E}_{\hat{z}\sim\pi_{\theta}(\cdot\mid P_S(q))}
\left[
\frac{1}{|\hat{z}|}
\sum_{k=1}^{|\hat{z}|}
D_{\mathrm{KL}}
\left(
p^S_k \,\|\, p^T_k
\right)
\right].
\label{eq:opd_loss}
\end{equation}
In practice, the divergence is computed only over the tool tokens and the end-of-sequence token.
Specifically, we truncate the student and teacher logits to this restricted output space and re-normalize them before computing the divergence~\cite{zhao2026self}, focusing distillation on executable tool-sequence generation rather than the full natural-language vocabulary.

In the second training stage, the student is optimized with the combined objective:
\begin{equation}
\mathcal{L}
=
\mathcal{L}_{\mathrm{sft}}
+
\lambda
\mathcal{L}_{\mathrm{opd}},
\label{eq:final_loss}
\end{equation}
where $\lambda$ controls the strength of on-policy distillation, and the SFT loss stabilizes policy optimization. Only the student policy is updated, while the teacher remains frozen. By doing so, the student is optimized under its own inference-time state distribution, while the teacher provides dense token-level supervision enriched with subtask reasoning knowledge.

\paragraph{Inference.}
At inference time, the model only takes the user query as input, without requiring explicit subtask decomposition, ground-truth trajectories, or additional graph context. Given only a user query $q$, the trained policy autoregressively generates a tool-token sequence
$\hat{z}=(\hat{z}_1,\ldots,\hat{z}_K)$ over the restricted tool-token vocabulary.
The sequence is then mapped back to the predicted tool trajectory $\hat{y}$ via the inverse token mapping.

The complete training procedure is summarized in Algorithm~\ref{alg:gtplan}, and the detailed student and teacher prompts are provided in Appendix~\ref{app:prompts}.

\section{Experiments}

\subsection{Experimental Settings}

\paragraph{Datasets.}
We evaluate \textsc{GRAFT} on four publicly available datasets with directed tool graphs: HuggingFace~\cite{wu2024can}, Multimedia~\cite{wu2024can}, UltraTool~\cite{huang2024planning}, and ToolBench~\cite{qin2023toolllm}. Specifically, HuggingFace and Multimedia are from TaskBench~\cite{shen2024taskbench}: HuggingFace~\cite{wu2024can} focuses on AI-task planning, while Multimedia~\cite{wu2024can} contains multimodal tool chains involving vision, audio, generation, and transformation tools. UltraTool~\cite{huang2024planning} features complex planning scenarios across thousands of tasks, covering travel, tourism, and other daily-life domains. ToolBench~\cite{qin2023toolllm} evaluates practical multi-step task scenarios in weather lookup, financial analysis, e-commerce queries, and social media information retrieval. 

The directed tool graphs in these datasets define valid successor relations among tools and each training sample contains a natural-language user query, ground-truth subtasks and ordered tool trajectory. The detailed information about graphs and datasets are summarized in the Appendix \ref{app_data}.

\begin{table*}[t]
\center

\caption{Comparison of our method with baselines under EM (\%), ELR, and ACPL. The best result is shown in \textbf{bold}, and the runner-up is \underline{underlined}.}

\label{tab_acc}
\renewcommand{\arraystretch}{1.15} 

\resizebox{\linewidth}{!}{
\begin{tabular}{l|l|ccc|ccc|ccc|ccc}
\toprule
\multicolumn{1}{c|}{\multirow{2}{*}{LLM}} & \multicolumn{1}{c|}{\multirow{2}{*}{Method}} & \multicolumn{3}{c|}{HuggingFace}                                            & \multicolumn{3}{c|}{Multimedia}                                             & \multicolumn{3}{c|}{UltraTool}                                              & \multicolumn{3}{c}{ToolBench}                                         \\ \cline{3-14} 
\multicolumn{1}{c|}{}                     & \multicolumn{1}{c|}{}                        & \multicolumn{1}{c}{EM} & \multicolumn{1}{c}{ELR} & \multicolumn{1}{c|}{ACPL} & \multicolumn{1}{c}{EM} & \multicolumn{1}{c}{ELR} & \multicolumn{1}{c|}{ACPL} & \multicolumn{1}{c}{EM} & \multicolumn{1}{c}{ELR} & \multicolumn{1}{c|}{ACPL} & \multicolumn{1}{c}{EM} & \multicolumn{1}{c}{ELR} & \multicolumn{1}{c}{ACPL} \\ \hline
\multirow{8}{*}{Qwen2.5-1.5B}             & BeamSearch                                   & 1.75                   & 0.92                    & 0.36                     & 3.52                   & 0.87                    & 0.41                     & 5.60                   & 0.63                    & 0.35                     & 10.59                  & {\ul 0.67}              & 0.51                    \\
                                          & Direct                                       & 4.85                   & 0.53                    & 0.44                     & 5.21                   & 0.45                    & 0.61                     & 3.62                   & 0.49                    & 0.34                     & 9.15                   & 0.31                    & 0.32                    \\
                                          & HuggingGPT                                   & 4.81                   & 0.48                    & 0.33                     & 5.41                   & 0.49                    & 0.67                     & 3.83                   & 0.45                    & 0.35                     & 9.21                   & 0.32                    & 0.36                    \\
                                          & PLaG                                         & 4.86                   & 0.49                    & 0.36                     & 5.21                   & 0.46                    & 0.67                     & 4.85                   & 0.46                    & 0.36                     & 15.29                  & 0.34                    & 0.65                    \\
                                          & GNN4Plan                                     & 13.43                  & \textbf{1.00}           & 0.99                     & 23.80                  & \textbf{1.00}           & 1.59                     & 12.02                  & \textbf{1.00}           & 0.65                     & 6.40                   & \textbf{1.00}           & 0.70                    \\
                                          & GTool                                        & 21.80                  & 0.78                    & 1.33                     & 48.00                  & 0.87                    & 1.67                     & 41.60                  & 0.78                    & 1.25                     & 10.80                  & 0.21                    & 0.51                    \\
                                          & ToolGen                                      & {\ul 50.00  }          & 0.91                    & {\ul 2.28}               & {\ul 55.20  }          & 0.91                    & {\ul 2.61}               & {\ul 53.00  }          & 0.89                    & {\ul 1.48}               & {\ul 31.40  }          & 0.28                    & {\ul 1.03}              \\
                                          & Ours                                         & \textbf{53.20  }       & {\ul 0.96}              & \textbf{2.33}            & \textbf{64.00  }       & {\ul 0.94}              & \textbf{2.75}            & \textbf{60.60  }       & {\ul 0.91}              & \textbf{1.63}            & \textbf{39.40  }       & 0.35                    & \textbf{1.28}           \\ \hline
\multirow{8}{*}{Llama-3.2-3B}             & BeamSearch                                   & 6.54                   & 0.93                    & 0.65                     & 5.58                   & 0.79                    & 0.54                     & 3.29                   & 0.62                    & 0.20                     & 11.70                  & {\ul 0.74}              & 0.57                    \\
                                          & Direct                                       & 8.69                   & 0.64                    & 0.71                     & 11.30                  & 0.76                    & 0.89                     & 4.90                   & 0.45                    & 0.37                     & 15.58                  & 0.22                    & 0.57                    \\
                                          & HuggingGPT                                   & 7.59                   & 0.66                    & 0.65                     & 11.51                  & 0.76                    & 0.88                     & 4.99                   & 0.46                    & 0.38                     & 15.15                  & 0.25                    & 0.57                    \\
                                          & PLaG                                         & 7.17                   & 0.67                    & 0.59                     & 12.84                  & 0.79                    & 0.88                     & 5.24                   & 0.48                    & 0.44                     & 16.86                  & 0.33                    & 0.67                    \\
                                          & GNN4Plan                                     & 18.40                  & \textbf{1.00}           & 1.20                     & 37.65                  & \textbf{1.00}           & 2.05                     & 16.84                  & \textbf{1.00}           & 0.74                     & 6.61                   & \textbf{1.00}           & 0.70                    \\
                                          & GTool                                        & 37.00                  & 0.91                    & 1.85                     & 56.80                  & 0.93                    & 2.55                     & 55.60                  & 0.90                    & 1.55                     & 11.80                  & 0.22                    & 0.52                    \\
                                          & ToolGen                                      & {\ul 57.20  }          & 0.96                    & {\ul 2.42}               & {\ul 67.80  }          & 0.91                    & {\ul 2.83}               & {\ul 58.60  }          & 0.92                    & {\ul 1.63}               & {\ul 33.00  }          & 0.31                    & {\ul 1.18}              \\
                                          & Ours                                         & \textbf{62.40  }       & {\ul 0.99}              & \textbf{2.62}            & \textbf{74.80  }       & {\ul 0.99}              & \textbf{3.09}            & \textbf{65.20  }       & {\ul 0.95}              & \textbf{1.75}            & \textbf{41.80  }       & 0.36                    & \textbf{1.33}           \\ \hline
\multirow{8}{*}{Mistral-7B}               & BeamSearch                                   & 12.53                  & 0.84                    & 0.93                     & 29.12                  & 0.91                    & 1.59                     & 15.45                  & 0.72                    & 0.65                     & 13.26                  & {\ul 0.51}              & 0.57                    \\
                                          & Direct                                       & 7.82                   & 0.67                    & 0.74                     & 16.06                  & 0.64                    & 1.10                     & 5.02                   & 0.47                    & 0.44                     & 15.47                  & 0.37                    & 0.52                    \\
                                          & HuggingGPT                                   & 7.39                   & 0.62                    & 0.53                     & 10.02                  & 0.55                    & 0.82                     & 8.38                   & 0.45                    & 0.40                     & 16.01                  & 0.38                    & 0.60                    \\
                                          & PLaG                                         & 8.53                   & 0.67                    & 0.56                     & 11.07                  & 0.62                    & 0.78                     & 10.52                  & 0.62                    & 0.48                     & 21.95                  & 0.39                    & 0.90                    \\
                                          & GNN4Plan                                     & 12.22                  & \textbf{1.00}           & 1.06                     & 34.27                  & \textbf{1.00}           & 1.95                     & 15.43                  & \textbf{1.00}           & 0.80                     & 6.70                   & \textbf{1.00}           & 0.72                    \\
                                          & GTool                                        & 45.20                  & 0.97                    & 2.04                     & 59.20                  & 0.94                    & 2.60                     & {\ul 63.80  }          & 0.93                    & 1.69                     & 13.80                  & 0.24                    & 0.58                    \\
                                          & ToolGen                                      & {\ul 60.00  }          & 0.98                    & {\ul 2.55}               & {\ul 68.40  }          & {\ul 0.99}              & {\ul 2.94}               & 62.00                  & 0.94                    & {\ul 1.71}               & {\ul 35.00  }          & 0.33                    & {\ul 1.21}              \\
                                          & Ours                                         & \textbf{63.20  }       & \textbf{1.00}           & \textbf{2.63}            & \textbf{74.20  }       & {\ul 0.99}              & \textbf{2.98}            & \textbf{66.20  }       & {\ul 0.96}              & \textbf{1.78}            & \textbf{42.60  }       & 0.40                    & \textbf{1.35}           \\ \bottomrule
\end{tabular}}
\end{table*}

\paragraph{Baselines.}

We compare \textsc{GRAFT} with seven representative baselines.
\textsc{BeamSearch}~\citep{wu2024can,10.1007/978-3-031-73254-6_6} performs constrained graph search with LLM-based scoring to generate candidate tool sequences.
\textsc{Direct} prompts the LLM with the user query and the tool pool, and directly asks it to generate the tool sequence.
\textsc{HuggingGPT}~\citep{shen2023hugginggpt} follows a step-by-step planning strategy that decomposes the user query into subtasks and assigns tools accordingly.
\textsc{PLaG}~\citep{pmlr-v235-lin24k} serializes the entire tool graph into plain text and incorporates it into the prompt to enhance LLM reasoning and tool planning.
\textsc{GNN4Plan}~\citep{wu2024can} encodes tool nodes with a graph neural network and uses the resulting node representations for subtask-level tool retrieval.
\textsc{GTool}~\citep{chen2026gtool} constructs a graph-aware token using a graph neural network and feeds it, together with the user query and tool pool, into the LLM for tool generation.
\textsc{ToolGen}~\citep{wang2025toolgen} represents tools as special tokens for generative tool retrieval, but does not model dependencies among tools.

We exclude agentic tool-use methods~\citep{wang2024llms,qian2024toolink,qin2023toolllm,li2025flow}, which rely on multi-turn tool execution and action feedback. These methods are orthogonal to the single turn execution in \textsc{GRAFT} and may further improve the performance, which we leave for future work.

\paragraph{Evaluation metrics.}
We evaluate planning correctness and dependency executability using three primary metrics.
Exact Match (EM) measures whether the predicted sequence exactly matches the ground-truth trajectory.
Edge Legality Rate (ELR) measures the fraction of predicted adjacent tool transitions that correspond to valid directed edges.
Average Correct Prefix Length (ACPL) measures the average number of correctly predicted tools before the first error.
We also report additional metrics, including Tool F1, Normalized Edit Distance, and Prefix Accuracy@K.
Their definitions and results are provided in Appendices~\ref{app_data} and~\ref{add_resu}, respectively.

\paragraph{Implementation details.}

We evaluate all methods with three open-source LLM backbones: Qwen2.5-1.5B-Instruct~\citep{yang2024qwen25}, Llama-3.2-3B-Instruct~\citep{meta2024llama32}, and Mistral-7B-Instruct~\citep{jiang2023mistral}. 
All models are fine-tuned with rank-16 LoRA~\citep{hu2021lora}.
To sample training sequences for edge reconstruction, the maximum path length $R_{\max}$ is set to the maximum tool-sequence length of each dataset, and path lengths follow the empirical length distribution.
We use a positive-to-negative edge ratio of $1{:}10$ and set the edge-scoring temperature to $\tau=0.1$.
The distillation loss weight is selected from $\{0,0.1,0.3,0.5,0.7,1,10\}$.
All experiments are conducted on a server with Intel Xeon Gold 6326 CPUs at 2.90GHz and NVIDIA A40 GPUs with 48GB memory.

\subsection{Main Results}
Table~\ref{tab_acc} compares all methods under EM, ELR, and ACPL. The first four baselines do not require task-specific fine-tuning, but show limited performance in exact sequence prediction. BeamSearch and GNN4Plan are graph search-based methods and thus achieve relatively high ELR. In particular, GNN4Plan strictly selects each next tool from graph-constrained neighbors, leading to an ELR of 1.00 in all cases. However, its much lower EM and ACPL suggest that local dependency validity alone is insufficient for accurate multi-step planning.
In contrast, our method consistently achieves the best EM and ACPL across all backbones and datasets, while maintaining competitive ELR. 
The higher ACPL indicates that our method preserves longer correct prefixes, thereby mitigating early planning failures in long tool sequences. Overall, these results demonstrate that our method effectively maintains tool dependency legality while substantially improving tool planning accuracy.

\subsection{Ablation Study}

Table~\ref{tab_ablation} reports the ablation results on Llama-3.2-3B. Here, \textit{w/o distill.} removes the on-policy distillation loss, \textit{w/o graph} removes edge reconstruction, \textit{w/o subtask} removes the subtask context from the teacher prompt, and \textit{w/o STT} removes subtask-to-tool grounding training. Removing distillation leads to the largest drop on both datasets, particularly in ACPL, indicating that on-policy learning is important for preserving longer correct prefixes. Removing graph reconstruction also degrades EM and ELR, confirming that learning directed tool dependencies improves both prediction accuracy and dependency legality. The performance drops from \textit{w/o subtask} and \textit{w/o STT} further show that subtask-level supervision helps align task decomposition with tool selection. Overall, the full model achieves the best results across all metrics, demonstrating that our proposed strategies provide complementary benefits.

\begin{wraptable}[10]{r}{0.50\linewidth}
\vspace{-20pt}
\centering
\caption{Ablation study on Llama-3.2-3B.}
\label{tab_ablation}
\vspace{-5pt}
\scriptsize
\setlength{\tabcolsep}{1.8pt}
\renewcommand{\arraystretch}{1.15}

\resizebox{\linewidth}{!}{
\begin{tabular}{@{}l|ccc|ccc@{}}
\toprule
\multirow{2}{*}{Method} 
& \multicolumn{3}{c|}{HuggingFace} 
& \multicolumn{3}{c}{Multimedia} \\ 
\cline{2-7}
& EM & ELR & ACPL & EM & ELR & ACPL \\ 
\hline
w/o distill. & 59.81 & 0.96 & 2.41 & 71.76 & 0.93 & 2.88 \\
w/o graph    & 60.25 & 0.96 & 2.44 & 72.90 & 0.94 & 2.90 \\
w/o subtask  & 61.90 & 0.98 & 2.57 & 73.23 & 0.96 & 2.93 \\
w/o STT      & 62.14 & 0.99 & 2.59 & 74.10 & 0.98 & 3.00 \\ 
\hline
Ours         & 62.40 & 0.99 & 2.62 & 74.80 & 0.99 & 3.09 \\ 
\bottomrule
\end{tabular}}
\vspace{0pt}
\end{wraptable}

\begin{figure*}[t]
\centering

\begin{subfigure}[t]{0.50\textwidth}
\centering
\vspace{0pt}
\includegraphics[width=\linewidth]{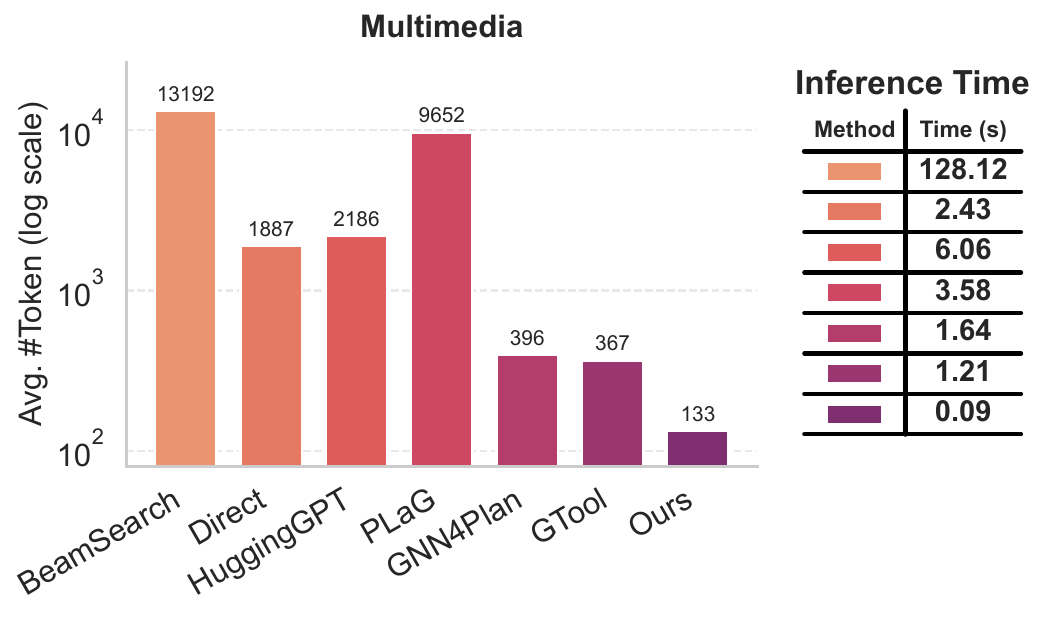}
\label{fig:efficiency}
\end{subfigure}
\hfill
\begin{subfigure}[t]{0.46\textwidth}
\centering
\vspace{0pt}
\includegraphics[width=\linewidth]{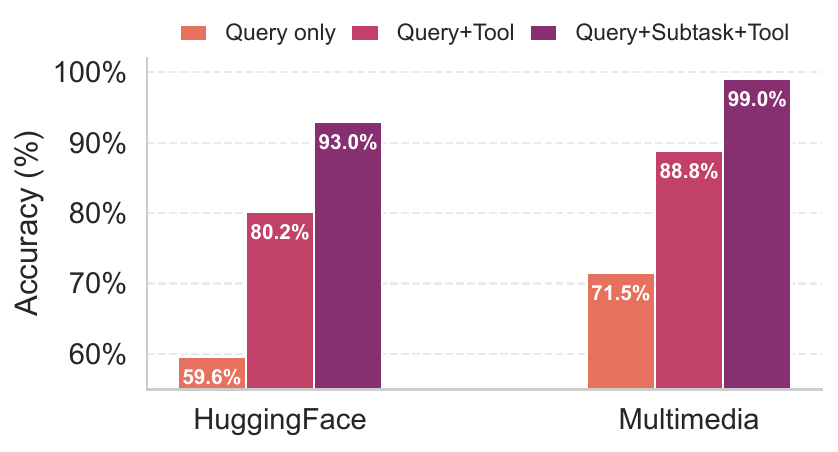}
\label{fig:teacher}
\end{subfigure}

\vspace{-5mm}
\caption{
Efficiency and teacher prompt analysis.
(a) compares total token cost and average inference time per test sample using Qwen2.5-1.5B.
(b) reports validation accuracy under different teacher prompt variants using Llama-3.2-3B.
}
\label{fig:analysis}
\vspace{-2mm}
\end{figure*}

\subsection{Efficiency Analysis}

We compare the token consumption and inference time of different methods in Figure~\ref{fig:analysis}(a). BeamSearch requires multiple iterative LLM calls, leading to the largest token cost and inference time. PLaG serializes the tool graph into the prompt, which substantially increases the input length and reasoning complexity. HuggingGPT also introduces additional subtask-level reasoning, resulting in relatively high token usage. GNN4Plan and GTool reduce token cost by encoding the graph with GNNs, but they still need to include the tool descriptions in the prompt. In contrast, our method internalizes tool graph knowledge through tokens and only requires the user query at inference time, achieving the lowest token consumption and significant reduction in per-query inference time.

\subsection{Teacher Prompt Analysis}

Figure~\ref{fig:analysis}(b) compares different teacher prompt designs to examine how planning context affects the quality of distillation supervision. Compared with the query-only prompt, providing the ground-truth tool sequence improves teacher prediction accuracy on both datasets. However, the accuracy is still far from perfect, indicating that the teacher does not simply copy the provided answer. Instead, the ground-truth sequence serves as useful planning guidance, while the teacher still needs to interpret the query and predict the next tool under the given context.
More importantly, adding subtask information leads to a substantial further improvement. This suggests that subtasks provide explicit intermediate intent and help the teacher better understand the stepwise structure of the task. 
Overall, these results confirm the effectiveness of subtask-aware teacher prompting and support our design of using privileged planning context to guide on-policy distillation.

\subsection{Hallucination Analysis}
\label{app_hall}

Figure~\ref{fig2all}(a) reports the hallucination ratio, defined as the proportion of generated tool names outside the tool pool among all generated tool predictions. Direct prompting yields the highest hallucination ratio, indicating that plain-text tool descriptions are insufficient for reliable tool grounding. BeamSearch reduces hallucination but still generates invalid tools, since its search space relies on both graph and LLM-estimated candidate tools. HuggingGPT, PLaG, and GTool also suffer from hallucination because they remain generative retrieval-based methods that select tools from prompt-provided descriptions. In contrast, GRAFT achieves a hallucination ratio of $0.00\%$. This is because each tool is represented as a dedicated token, restricting generation to the tool-token space, while subtask grounding and query-to-tool SFT further align tool tokens with practical usage and task-level planning.

\subsection{Hyperparameter Sensitivity}
\label{app_hyperpara}

\begin{figure*}[t]
\centering

\begin{subfigure}[t]{0.62\textwidth}
\centering
\vspace{0pt}
\includegraphics[width=\linewidth]{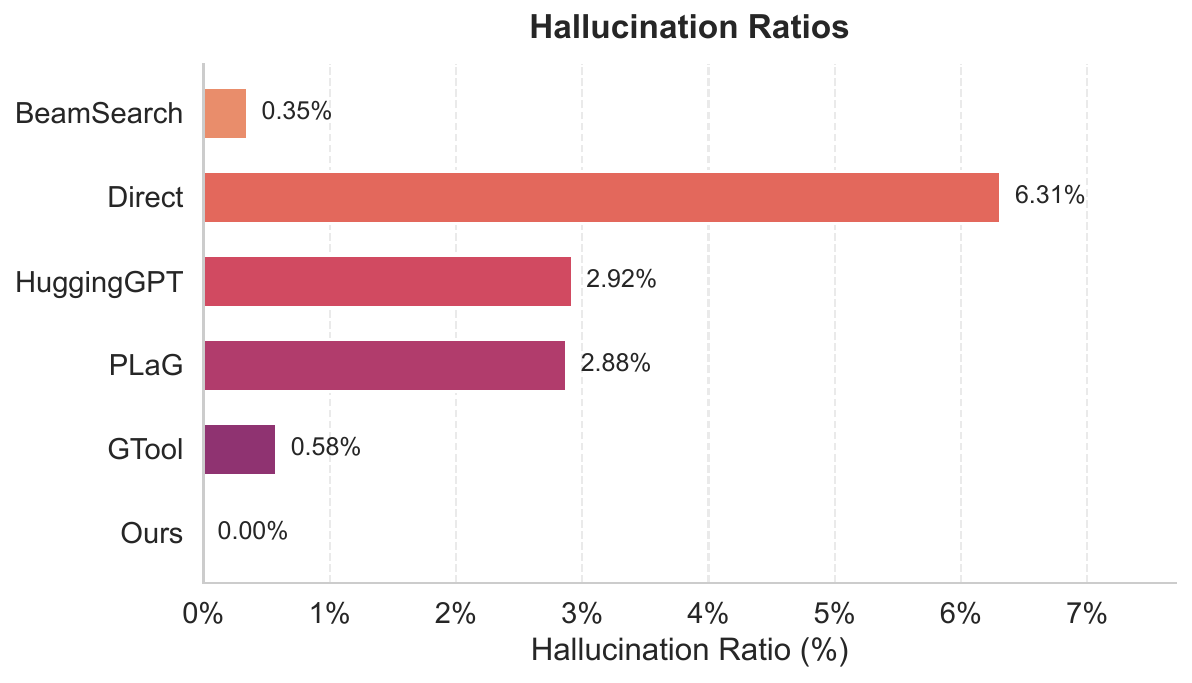}
\end{subfigure}
\hfill
\begin{subfigure}[t]{0.37\textwidth}
\centering
\vspace{0pt}
\includegraphics[width=\linewidth]{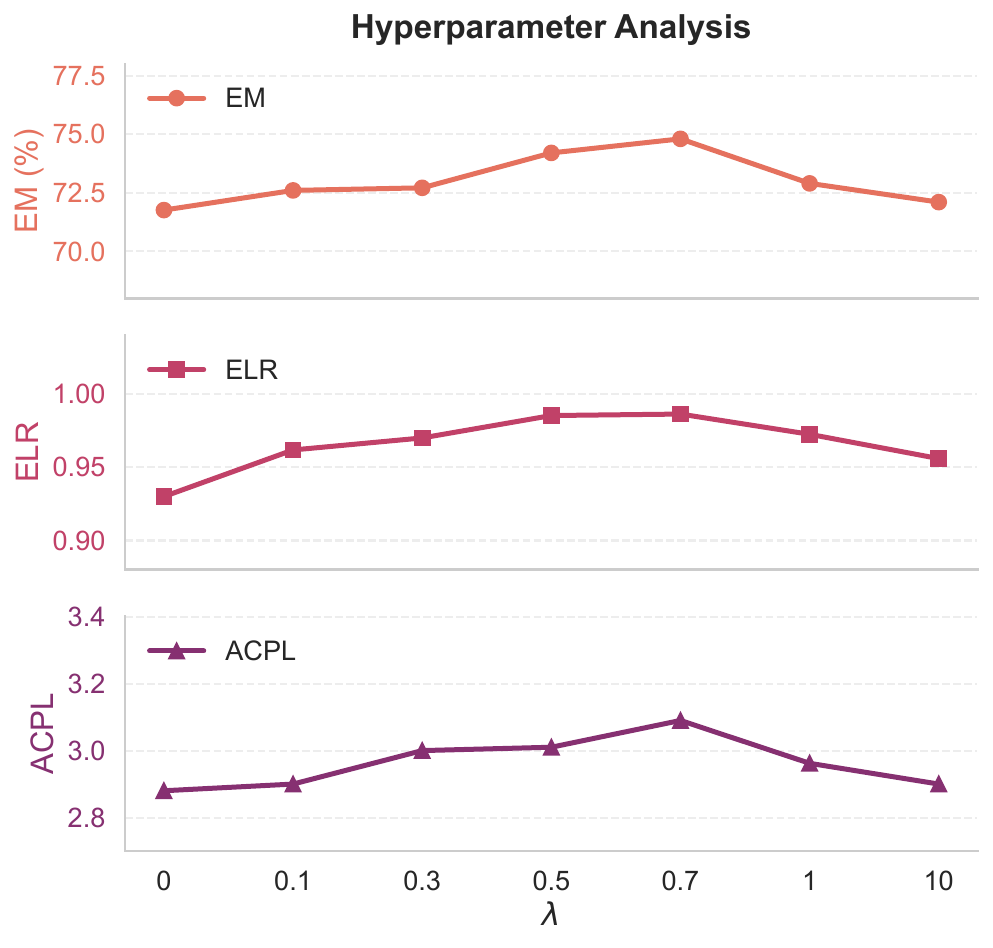}
\end{subfigure}

\caption{
Hallucination analysis and hyperparameter sensitivity analysis of different methods using Llama-3.2-3B on the Multimedia dataset.
}
\label{fig2all}
\vspace{-2mm}
\end{figure*}

Figure~\ref{fig2all} (b) analyzes the sensitivity to the distillation weight $\lambda$ on the Multimedia dataset with Llama-3.2-3B. As $\lambda$ increases from $0$ to $0.7$, all three metrics improve consistently, indicating that on-policy distillation benefits both task-level tool selection and dependency-aware sequence generation. However, performance declines when $\lambda$ becomes too large, suggesting that an overly strong soft distillation loss may dominate the SFT objective and weaken the model's alignment with ground-truth tool trajectories. Overall, moderate distillation is most effective: both insufficient and excessive distillation lead to suboptimal performance.

\section{Conclusion}
\label{sec_conclusion}

We presented \textsc{GRAFT}, a framework for multi-step tool calling that internalizes directed tool dependencies through graph-token learning and on-policy tool context distillation. By reconstructing successor edges from tool-token hidden states and learning from student-generated trajectories under a subtask-aware privileged teacher, \textsc{GRAFT} significantly improves both exact planning accuracy and dependency legality than baselines on multiple real-world benchmarks.

A potential limitation is the lack of tool-action execution feedback. Future work will extend \textsc{GRAFT} to interactive agentic environments with tool dependencies, where planning can be refined through tool observations and execution outcomes.

{\small
\bibliographystyle{unsrt}
\bibliography{ref}}

@article{ye2026policy,
  title={On-policy context distillation for language models},
  author={Ye, Tianzhu and Dong, Li and Wu, Xun and Huang, Shaohan and Wei, Furu},
  journal={arXiv preprint arXiv:2602.12275},
  year={2026}
}

@article{zhao2026self,
  title={Self-Distilled Reasoner: On-Policy Self-Distillation for Large Language Models},
  author={Zhao, Siyan and Xie, Zhihui and Liu, Mengchen and Huang, Jing and Pang, Guan and Chen, Feiyu and Grover, Aditya},
  journal={arXiv preprint arXiv:2601.18734},
  year={2026}
}

@article{li2025flow,
  title={In-the-flow agentic system optimization for effective planning and tool use},
  author={Li, Zhuofeng and Zhang, Haoxiang and Han, Seungju and Liu, Sheng and Xie, Jianwen and Zhang, Yu and Choi, Yejin and Zou, James and Lu, Pan},
  journal={arXiv preprint arXiv:2510.05592},
  year={2025}
}

@article{xi2025agentgym,
  title={Agentgym-rl: Training llm agents for long-horizon decision making through multi-turn reinforcement learning},
  author={Xi, Zhiheng and Huang, Jixuan and Liao, Chenyang and Huang, Baodai and Guo, Honglin and Liu, Jiaqi and Zheng, Rui and Ye, Junjie and Zhang, Jiazheng and Chen, Wenxiang and others},
  journal={arXiv preprint arXiv:2509.08755},
  year={2025}
}

@article{yang2024qwen25,
  title={Qwen2.5 Technical Report},
  author={Yang, An and Yang, Baosong and Zhang, Beichen and Hui, Binyuan and Zheng, Bo and Yu, Bowen and others},
  journal={arXiv preprint arXiv:2412.15115},
  year={2024}
}

@inproceedings{huang2024planning,
  title={Planning, creation, usage: Benchmarking llms for comprehensive tool utilization in real-world complex scenarios},
  author={Huang, Shijue and Zhong, Wanjun and Lu, Jianqiao and Zhu, Qi and Gao, Jiahui and Liu, Weiwen and Hou, Yutai and Zeng, Xingshan and Wang, Yasheng and Shang, Lifeng and others},
  booktitle={Findings of the Association for Computational Linguistics: ACL 2024},
  pages={4363--4400},
  year={2024}
}

@misc{meta2024llama32,
  title={Llama 3.2 3B Instruct Model Card},
  author={{Meta AI}},
  year={2024},
  howpublished={\url{https://huggingface.co/meta-llama/Llama-3.2-3B-Instruct}},
  note={Accessed: 2026-05-07}
}

@article{jiang2023mistral,
  title={Mistral 7B},
  author={Jiang, Albert Q. and Sablayrolles, Alexandre and Mensch, Arthur and Bamford, Chris and Chaplot, Devendra Singh and de las Casas, Diego and Bressand, Florian and Lengyel, Gianna and Lample, Guillaume and Saulnier, Lucile and others},
  journal={arXiv preprint arXiv:2310.06825},
  year={2023}
}

@inproceedings{wang2024llms,
  title={Llms in the imaginarium: tool learning through simulated trial and error},
  author={Wang, Boshi and Fang, Hao and Eisner, Jason and Van Durme, Benjamin and Su, Yu},
  booktitle={Proceedings of the 62nd Annual Meeting of the Association for Computational Linguistics (Volume 1: Long Papers)},
  pages={10583--10604},
  year={2024}
}

@inproceedings{zhang2026toward,
  title={Toward Graph-Tokenizing Large Language Models with Reconstructive Graph Instruction Tuning},
  author={Zhang, Zhongjian and Wang, Xiao and Zhang, Mengmei and Tan, Jiarui and Shi, Chuan},
  booktitle={Proceedings of the ACM Web Conference 2026},
  pages={430--441},
  year={2026}
}

@article{jin2024large,
  title={Large language models on graphs: A comprehensive survey},
  author={Jin, Bowen and Liu, Gang and Han, Chi and Jiang, Meng and Ji, Heng and Han, Jiawei},
  journal={IEEE Transactions on Knowledge and Data Engineering},
  volume={36},
  number={12},
  pages={8622--8642},
  year={2024},
  publisher={IEEE}
}

@inproceedings{gu2024minillm,
  title={Minillm: Knowledge distillation of large language models},
  author={Gu, Yuxian and Dong, Li and Wei, Furu and Huang, Minlie},
  booktitle={The twelfth international conference on learning representations},
  year={2024}
}

@article{song2026survey,
  title={A Survey of On-Policy Distillation for Large Language Models},
  author={Song, Mingyang and Zheng, Mao},
  journal={arXiv preprint arXiv:2604.00626},
  year={2026}
}

@article{li2023survey,
  title={A survey of graph meets large language model: Progress and future directions},
  author={Li, Yuhan and Li, Zhixun and Wang, Peisong and Li, Jia and Sun, Xiangguo and Cheng, Hong and Yu, Jeffrey Xu},
  journal={arXiv preprint arXiv:2311.12399},
  year={2023}
}

@misc{anonymous2026graphtoolbench,
  title={GraphToolBench: Benchmarking LLMs for Sequential Graph Comprehension and Conflict Identification in Tool Learning},
  author={Anonymous},
  year={2026},
  note={OpenReview},
  url={https://openreview.net/forum?id=zABFKZKtjK}
}

@article{masterman2024landscape,
  title={The landscape of emerging ai agent architectures for reasoning, planning, and tool calling: A survey},
  author={Masterman, Tula and Besen, Sandi and Sawtell, Mason and Chao, Alex},
  journal={arXiv preprint arXiv:2404.11584},
  year={2024}
}

@article{yao2022react,
  title={React: Synergizing reasoning and acting in language models},
  author={Yao, Shunyu and Zhao, Jeffrey and Yu, Dian and Du, Nan and Shafran, Izhak and Narasimhan, Karthik and Cao, Yuan},
  journal={arXiv preprint arXiv:2210.03629},
  year={2022}
}

@article{shen2024taskbench,
  title={Taskbench: Benchmarking large language models for task automation},
  author={Shen, Yongliang and Song, Kaitao and Tan, Xu and Zhang, Wenqi and Ren, Kan and Yuan, Siyu and Lu, Weiming and Li, Dongsheng and Zhuang, Yueting},
  journal={Advances in Neural Information Processing Systems},
  volume={37},
  pages={4540--4574},
  year={2024}
}

@article{shen2023hugginggpt,
  title={Hugginggpt: Solving ai tasks with chatgpt and its friends in hugging face},
  author={Shen, Yongliang and Song, Kaitao and Tan, Xu and Li, Dongsheng and Lu, Weiming and Zhuang, Yueting},
  journal={Advances in Neural Information Processing Systems},
  volume={36},
  pages={38154--38180},
  year={2023}
}

@article{wu2024can,
  title={Can graph learning improve planning in llm-based agents?},
  author={Wu, Xixi and Shen, Yifei and Shan, Caihua and Song, Kaitao and Wang, Siwei and Zhang, Bohang and Feng, Jiarui and Cheng, Hong and Chen, Wei and Xiong, Yun and others},
  journal={Advances in Neural Information Processing Systems},
  volume={37},
  pages={5338--5383},
  year={2024}
}

@inproceedings{
chen2026gtool,
title={{GT}ool: Graph Enhanced Tool Planning with Large Language Model},
author={Wenjie Chen and Di Yao and Wenbin Li and Xuying Meng and Chang Gong and Jingping Bi},
booktitle={The Fourteenth International Conference on Learning Representations},
year={2026},
url={https://openreview.net/forum?id=bn47cqGQ7l}
}

@inproceedings{
wang2025toolgen,
title={ToolGen: Unified Tool Retrieval and Calling via Generation},
author={Renxi Wang and Xudong Han and Lei Ji and Shu Wang and Timothy Baldwin and Haonan Li},
booktitle={The Thirteenth International Conference on Learning Representations},
year={2025},
url={https://openreview.net/forum?id=XLMAMmowdY}
}

@InProceedings{pmlr-v235-lin24k,
  title={Graph-enhanced Large Language Models in Asynchronous Plan Reasoning},
  author={Lin, Fangru and La Malfa, Emanuele and Hofmann, Valentin and Yang, Elle Michelle and Cohn, Anthony G. and Pierrehumbert, Janet B.},
  booktitle={Proceedings of the 41st International Conference on Machine Learning},
  pages={30108--30134},
  year={2024},
  editor={Salakhutdinov, Ruslan and Kolter, Zico and Heller, Katherine and Weller, Adrian and Oliver, Nuria and Scarlett, Jonathan and Berkenkamp, Felix},
  volume={235},
  series={Proceedings of Machine Learning Research},
  month={21--27 Jul},
  publisher={PMLR},
  url={https://proceedings.mlr.press/v235/lin24k.html}
}

@InProceedings{10.1007/978-3-031-73254-6_6,
author="Liu, Zhaoyang
and Lai, Zeqiang
and Gao, Zhangwei
and Cui, Erfei
and Li, Ziheng
and Zhu, Xizhou
and Lu, Lewei
and Chen, Qifeng
and Qiao, Yu
and Dai, Jifeng
and Wang, Wenhai",
editor="Leonardis, Ale{\v{s}}
and Ricci, Elisa
and Roth, Stefan
and Russakovsky, Olga
and Sattler, Torsten
and Varol, G{\"u}l",
title="ControlLLM: Augment Language Models with Tools by Searching on Graphs",
booktitle="Computer Vision -- ECCV 2024",
year="2025",
publisher="Springer Nature Switzerland",
address="Cham",
pages="89--105",
isbn="978-3-031-73254-6"
}

@inproceedings{qin2023toolllm,
  title={Toolllm: Facilitating large language models to master 16000+ real-world apis},
  author={Qin, Yujia and Liang, Shihao and Ye, Yining and Zhu, Kunlun and Yan, Lan and Lu, Yaxi and Lin, Yankai and Cong, Xin and Tang, Xiangru and Qian, Bill and others},
  booktitle={The twelfth international conference on learning representations},
  year={2023}
}

@inproceedings{qian2024toolink,
  title={Toolink: Linking toolkit creation and using through chain-of-solving on open-source model},
  author={Qian, Cheng and Xiong, Chenyan and Liu, Zhenghao and Liu, Zhiyuan},
  booktitle={Proceedings of the 2024 Conference of the North American Chapter of the Association for Computational Linguistics: Human Language Technologies (Volume 1: Long Papers)},
  pages={831--854},
  year={2024}
}

@article{yang2023gpt4tools,
  title={GPT4Tools: Teaching Large Language Model to Use Tools via Self-instruction},
  author={Yang, Rui and Song, Lin and Li, Yanwei and Zhao, Sijie and Ge, Yixiao and Li, Xiu and Shan, Ying},
  journal={arXiv preprint arXiv:2305.18752},
  year={2023}
}

@inproceedings{ross2011dagger,
  title={A Reduction of Imitation Learning and Structured Prediction to No-Regret Online Learning},
  author={Ross, Stephane and Gordon, Geoffrey J. and Bagnell, J. Andrew},
  booktitle={Proceedings of the Fourteenth International Conference on Artificial Intelligence and Statistics},
  pages={627--635},
  year={2011}
}

@article{hu2021lora,
  title={LoRA: Low-Rank Adaptation of Large Language Models},
  author={Hu, Edward J. and Shen, Yelong and Wallis, Phillip and Allen-Zhu, Zeyuan and Li, Yuanzhi and Wang, Shean and Wang, Lu and Chen, Weizhu},
  journal={arXiv preprint arXiv:2106.09685},
  year={2021}
}

@article{williams1992reinforce,
  title={Simple Statistical Gradient-Following Algorithms for Connectionist Reinforcement Learning},
  author={Williams, Ronald J.},
  journal={Machine Learning},
  volume={8},
  pages={229--256},
  year={1992}
}

@inproceedings{schulman2015trpo,
  title={Trust Region Policy Optimization},
  author={Schulman, John and Levine, Sergey and Moritz, Philipp and Jordan, Michael I. and Abbeel, Pieter},
  booktitle={Proceedings of the 32nd International Conference on Machine Learning},
  pages={1889--1897},
  year={2015}
}

@article{schulman2017ppo,
  title={Proximal Policy Optimization Algorithms},
  author={Schulman, John and Wolski, Filip and Dhariwal, Prafulla and Radford, Alec and Klimov, Oleg},
  journal={arXiv preprint arXiv:1707.06347},
  year={2017}
}

@inproceedings{bengio2015scheduled,
  title={Scheduled Sampling for Sequence Prediction with Recurrent Neural Networks},
  author={Bengio, Samy and Vinyals, Oriol and Jaitly, Navdeep and Shazeer, Noam},
  booktitle={Advances in Neural Information Processing Systems},
  year={2015}
}

@inproceedings{ranzato2016sequence,
  title={Sequence Level Training with Recurrent Neural Networks},
  author={Ranzato, Marc'Aurelio and Chopra, Sumit and Auli, Michael and Zaremba, Wojciech},
  booktitle={International Conference on Learning Representations},
  year={2016}
}

@inproceedings{lamb2016professor,
  title={Professor Forcing: A New Algorithm for Training Recurrent Networks},
  author={Lamb, Alex and Goyal, Anirudh and Zhang, Ying and Zhang, Saizheng and Courville, Aaron and Bengio, Yoshua},
  booktitle={Advances in Neural Information Processing Systems},
  year={2016}
}

@article{xu2025llm,
  title={LLM-Based Agents for Tool Learning: A Survey: W. Xu et al.},
  author={Xu, Weikai and Huang, Chengrui and Gao, Shen and Shang, Shuo},
  journal={Data Science and Engineering},
  pages={1--31},
  year={2025},
  publisher={Springer}
}

@article{shen2024llm,
  title={Llm with tools: A survey},
  author={Shen, Zhuocheng},
  journal={arXiv preprint arXiv:2409.18807},
  year={2024}
}

@article{liu2024toolnet,
  title={Toolnet: Connecting large language models with massive tools via tool graph},
  author={Liu, Xukun and Peng, Zhiyuan and Yi, Xiaoyuan and Xie, Xing and Xiang, Lirong and Liu, Yuchen and Xu, Dongkuan},
  journal={arXiv preprint arXiv:2403.00839},
  year={2024}
}

\newpage
\appendix

\begin{center}
\Large \bf {Appendix}
\end{center}

\etocdepthtag.toc{mtappendix}
\etocsettagdepth{mtchapter}{none}
\etocsettagdepth{mtappendix}{subsection}
\tableofcontents
\newpage



\section{Algorithm}
\label{alg:gtplan}

Algorithm~\ref{alg:gtplan2} outlines the optimization pipeline of GRAFT.
The first stage constructs graph-tokenized tool representations and grounds them with subtask-level supervision.
The second stage samples directed tool paths from $\mathcal{G}$ and optimizes edge reconstruction to internalize tool dependencies.
The third stage warms up the planner with query-to-tool SFT.
The final stage performs on-policy distillation, where the student is updated with the combined SFT and distillation objective under its own sampled trajectories.

\begin{algorithm}[ht]
\SetAlgoVlined
\small
\textbf{Input:} Training data $\mathcal{D}=\{(q,s,y)\}$, tool graph $\mathcal{G}=(\mathcal{T},\mathcal{E})$, LLM $\pi_{\theta}$ \\
\textbf{Output:} Trained planner $\pi_{\theta}$ \\

Map tools to dedicated tokens by Eq.~(\ref{eq:tool_token_vocab}); \\

Construct $\mathcal{D}_{\mathrm{sub}}$ by Eq.~(\ref{eq:subtask_pairs}) and update $\pi_{\theta}$ with $\mathcal{L}_{\mathrm{sub}}$ in Eq.~(\ref{eq:subtask_loss})
\tcp*{Tokenization \& Subtask SFT}


Construct a path set $\mathcal{R}$ by sampling $\rho$ from $\mathcal{G}$ via Eq.~(\ref{eq:graph_walk}); \\

\For{$epoch=1,\ldots,E_{\mathrm{edge}}$}{
    \ForEach{mini-batch $\mathcal{B}\subset\mathcal{R}$}{
        Compute successor scores by Eq.~(\ref{eq:successor_score}); \\
        Update $\pi_{\theta}$ with $\mathcal{L}_{\mathrm{edge}}$ in Eq.~(\ref{eq:edge_loss})
        \tcp*{Graph Reconstruction}
    }
}

\For{$epoch=1,\ldots,E_{\mathrm{sft}}$}{
    \ForEach{mini-batch $\mathcal{B}\subset\mathcal{D}$}{
        Compute $\mathcal{L}_{\mathrm{sft}}$ on $\mathcal{B}$ by Eq.~(\ref{eq:sft_loss}); \\
        Update $\pi_{\theta}$ with $\mathcal{L}_{\mathrm{sft}}$
        \tcp*{Query-to-tool Warm-up}

    }
}

Initialize student $\pi_{\theta}$ and frozen teacher $\pi_{\bar{\theta}}$ from the SFT checkpoint; \\

\For{$epoch=1,\ldots,E_{\mathrm{distill}}$}{
    \ForEach{mini-batch $\mathcal{B}\subset\mathcal{D}$}{
        Sample student trajectories $\hat{z}\sim\pi_{\theta}(\cdot\mid P_S(q))$; \\
        Compute $\mathcal{L}_{\mathrm{sft}}$ by Eq.~(\ref{eq:sft_loss}) and $\mathcal{L}_{\mathrm{opd}}$ by Eq.~(\ref{eq:opd_loss}); \\
        Update $\pi_{\theta}$ with $\mathcal{L}=\mathcal{L}_{\mathrm{sft}}+\lambda\mathcal{L}_{\mathrm{opd}}$ in Eq.~(\ref{eq:final_loss}) 
        \tcp*{Distillation + SFT}

    }
}
\textbf{Return:} Optimized $\pi_{\theta}$ \\

\caption{Overall optimization pipeline of GRAFT.}
\label{alg:gtplan2}
\end{algorithm}

\section{Datasets}
\label{app_data}


We evaluate \textsc{GRAFT} on four publicly available datasets with directed tool graphs: HuggingFace~\cite{wu2024can}, Multimedia~\cite{wu2024can}, UltraTool~\cite{huang2024planning}, and ToolBench~\cite{qin2023toolllm}.
To ensure consistent evaluation across different benchmarks, we process all datasets into a unified sequential tool-planning format. In the original graph-structured annotations, nodes represent tools and directed edges represent dependency relations between tools. During preprocessing, we use these original task nodes and graph edges to recover the ordered tool invocation sequence. After reconstruction, each final dataset instance contains the sample ID, user query, decomposed subtasks, and the ordered tool sequence. Fig. (\ref{dataset_examples}) shows representative examples from the experimental datasets.

For HuggingFace, Multimedia, and UltraTool, we use the datasets adopted in GNN4Plan~\cite{wu2024can} as our base datasets and retain the original tool pool and tool graph of each dataset. We first reformat the original data into a unified sequential structure and retain multi-tool invocation samples with chain-structured dependencies. For each retained sample, we recover the tool invocation path from its original dependency annotations and extract the ordered tool sequence associated with the given user query as the ground truth. 

As for ToolBench, we follow the GraphToolBench~\cite{anonymous2026graphtoolbench} and use the publicly released data construction methodology and code. 
Specifically, 612 high quality tools are selected and GPT 5.4 is used to generate the dependency-aware user query, together with intermediate task nodes and dependency links. These intermediate annotations are used to recover the ordered tool invocation sequence. 

\begin{figure*}[t]
\centering
\includegraphics[width=\linewidth]{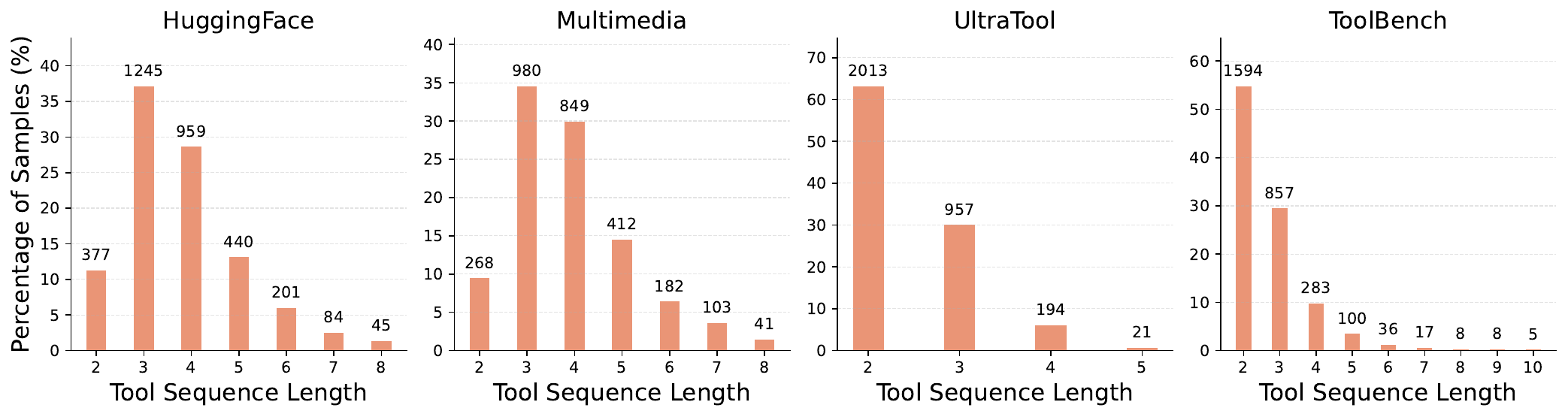}
\caption{The distribution of tool sequence length for different datasets.}
\label{seq_fig}
\end{figure*}

\begin{figure*}[t]
\centering
\includegraphics[width=\linewidth]{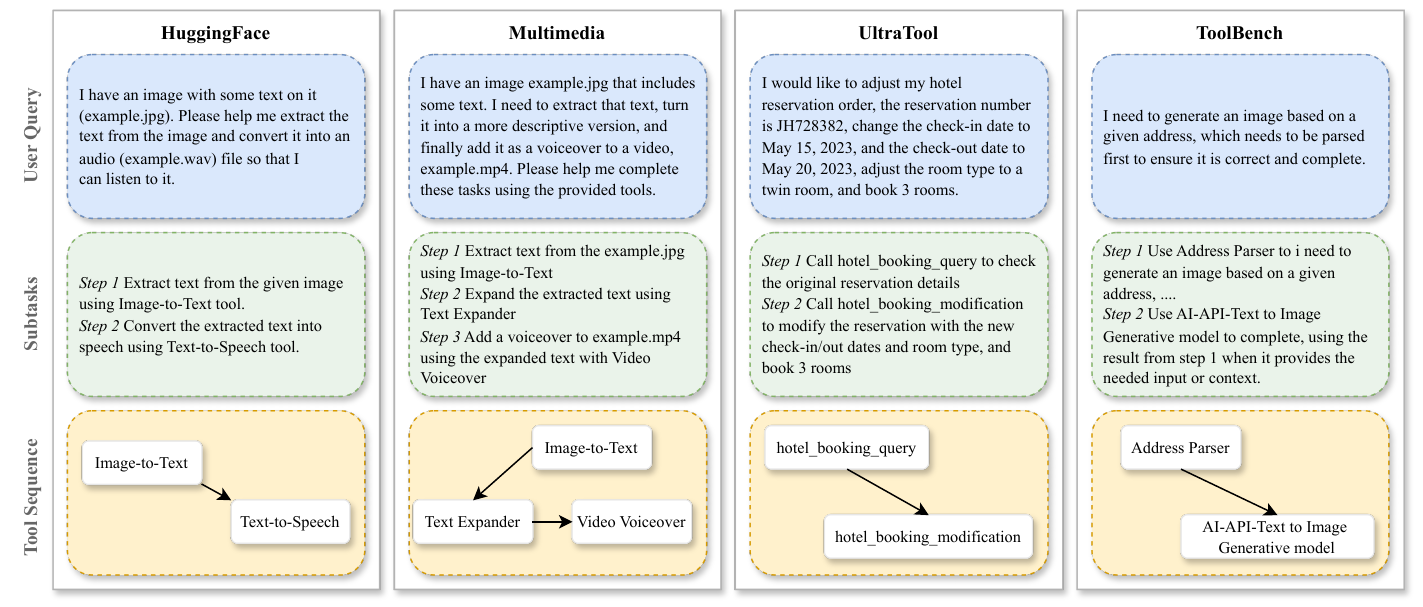}
\caption{Representative examples from the experimental datasets.}
\label{dataset_examples}
\end{figure*}

Each dataset is randomly split into training, validation, and test sets, with 500 samples for validation, 500 samples for testing, and the remaining samples for training. The detailed statistic of dataset and the tool sequence length distribution are shown in Table \ref{tab:dataset_statistics} and Fig. (\ref{seq_fig}), respectively.

\begin{table}[t]
\centering
\small
\caption{Statistics of the evaluated datasets. ``Max Len.'' denotes the maximum length of the ground truth tool sequence.}
\label{tab:dataset_statistics}
\begin{tabular}{lrrrrrrr}
\toprule
\textbf{Dataset} 
& \textbf{\#Tools} 
& \textbf{\#Edges} 
& \textbf{\#Train} 
& \textbf{\#Val} 
& \textbf{\#Test} 
& \textbf{Max Len.} \\
\midrule
HuggingFace & 23  & 225  & 2351 & 500 & 500 & 8   \\
Multimedia  & 40  & 449  & 1835 & 500 & 500 & 8   \\
UltraTool & 260 & 611  & 2185 & 500 & 500 & 5   \\
ToolBench   & 612 & 3140 & 1908 & 500 & 500 & 10  \\
\bottomrule
\end{tabular}
\end{table}

\section{Evaluation Metrics}
\label{app_metrics}
\paragraph{Evaluation metrics.}
We report the following metrics to evaluate our proposed method.

\paragraph{Exact Match (EM).}
Exact Match measures whether the predicted tool sequence exactly matches the ground-truth trajectory:
\begin{equation}
\mathrm{EM}
=
\frac{1}{N}
\sum_{i=1}^{N}
\mathbb{I}\left[\hat{y}_i = y_i\right].
\end{equation}
EM is the strictest metric, requiring both tool identities and their order to be fully correct.

\paragraph{Edge Legality Rate (ELR).}
Edge Legality Rate measures whether predicted adjacent tool transitions satisfy the directed dependency constraints in the tool graph.
For each example, we compute
\begin{equation}
\mathrm{ELR}_i
=\frac{1}{\hat{L}_i-1}
\sum_{k=1}^{\hat{L}_i-1}
\mathbb{I}\left[
(\hat{y}_{i,k},\hat{y}_{i,k+1})\in\mathcal{E}
\right],
\end{equation}
The final score is averaged over all test examples:
$
\mathrm{ELR}
=
\frac{1}{N}
\sum_{i=1}^{N}
\mathrm{ELR}_i
$.
ELR evaluates whether the generated trajectory is executable under the tool graph, rather than whether it exactly matches the reference.

\paragraph{Average Correct Prefix Length (ACPL).}
Average Correct Prefix Length measures the number of correctly predicted tools before the first mismatch:
\begin{equation}
c_i
=
\max
\left\{
k:
0\leq k\leq \min(L_i,\hat{L}_i),
\;
\hat{y}_{i,1:k}=y_{i,1:k}
\right\},
\qquad
\mathrm{ACPL}
=
\frac{1}{N}
\sum_{i=1}^{N}
c_i .
\end{equation}
ACPL reflects how far the model can follow the ground-truth trajectory before making its first error.

\paragraph{Tool F1.}
Tool F1 evaluates tool selection accuracy regardless of order.
For each example, we compare the set of predicted tools with the set of ground-truth tools, and compute the harmonic mean of node precision and node recall.
This metric captures whether the model chooses the correct tools, but ignores their ordering.

\paragraph{Normalized Edit Distance (NED).}
Normalized Edit Distance measures the sequence-level distance between prediction and ground truth.
We compute the Levenshtein edit distance with unit insertion, deletion, and substitution costs, and normalize it by the maximum sequence length:
\begin{equation}
\mathrm{NED}_i
=
\frac{
\mathrm{EditDist}(\hat{y}_i,y_i)
}{
\max(\hat{L}_i,L_i)
}.
\end{equation}
A lower NED indicates a prediction closer to the ground-truth trajectory.

\paragraph{Prefix Accuracy@K (PA@K).}
Prefix Accuracy@K measures whether the model correctly predicts the first $K$ tools of the ground-truth trajectory.
Using the correct prefix length $c_i$, it is computed as
\begin{equation}
\mathrm{PA@}K
=
\frac{1}{|\mathcal{D}_K|}
\sum_{i\in\mathcal{D}_K}
\mathbb{I}[c_i \geq K],
\qquad
\mathcal{D}_K=\{i\mid L_i\geq K\}.
\end{equation}
This metric evaluates prefix correctness at different planning depths.

\section{Additional Results}
\label{add_resu}

Table~\ref{tab_acc2} reports the comparison under Tool F1, NED, PA@1, and PA@3.
Tool F1 measures whether the predicted tool set matches the required tools, reflecting tool selection quality.
NED evaluates the normalized edit distance between the predicted and ground-truth tool sequences, where a lower value indicates better sequence-level alignment.
PA@1 and PA@3 measure whether the generated plan matches the ground-truth sequence within the first one and first three steps, respectively.
Across all methods, PA@1 is consistently higher than PA@3, indicating that predicting the first tool is relatively easier, while maintaining correctness over later tool steps becomes more challenging due to the increasing difficulty of long-horizon dependency modeling.
Across both datasets and all three LLM backbones, GRAFT consistently achieves the best performance on all metrics.
Compared with the strongest baseline ToolGen, GRAFT further improves Tool F1 and PA while reducing NED, showing that graph-tokenized dependency learning and on-policy distillation improve both tool selection and sequence-level planning accuracy.

\begin{table*}[t]
\center

\caption{Comparison of our method with baselines under Tool F1 (\%), NED, PA@1 and PA@3. The best result is shown in \textbf{bold}.}

\label{tab_acc2}
\renewcommand{\arraystretch}{1.15} 

\resizebox{\linewidth}{!}{
\begin{tabular}{l|l|cccc|cccc}
\hline
\multicolumn{1}{c|}{\multirow{2}{*}{LLM}} & \multicolumn{1}{c|}{\multirow{2}{*}{Method}} & \multicolumn{4}{c|}{HuggingFace}                                 & \multicolumn{4}{c}{Multimedia}                                   \\ \cline{3-10} 
\multicolumn{1}{c|}{}                     & \multicolumn{1}{c|}{}                        & Tool F1          & NED           & PA@1          & PA@3          & Tool F1          & NED           & PA@1          & PA@3          \\ \hline
\multirow{8}{*}{Qwen2.5-1.5B}             & BeamSearch                                   & 35.59            & 0.76          & 0.28          & 0.02          & 37.72            & 0.71          & 0.22          & 0.06          \\
                                          & Direct                                       & 38.21            & 0.70          & 0.28          & 0.05          & 42.99            & 0.65          & 0.37          & 0.07          \\
                                          & HuggingGPT                                   & 32.68            & 0.74          & 0.20          & 0.03          & 44.62            & 0.63          & 0.38          & 0.09          \\
                                          & PLaG                                         & 29.92            & 0.76          & 0.21          & 0.05          & 43.24            & 0.66          & 0.35          & 0.07          \\
                                          & GNN4Plan                                     & 58.82            & 0.52          & 0.47          & 0.16          & 68.13            & 0.42          & 0.59          & 0.36          \\
                                          & GTool                                        & 67.21            & 0.43          & 0.64          & 0.25          & 77.39            & 0.24          & 0.60          & 0.49          \\
                                          & ToolGen                                      & 85.98            & 0.19          & 0.78          & 0.54          & 88.36            & 0.18          & 0.84          & 0.62          \\
                                          & Ours                                         & \textbf{86.50  } & \textbf{0.18} & \textbf{0.79} & \textbf{0.57} & \textbf{90.52  } & \textbf{0.16} & \textbf{0.85} & \textbf{0.66} \\ \hline
\multirow{8}{*}{Llama-3.2-3B}             & BeamSearch                                   & 49.18            & 0.62          & 0.38          & 0.08          & 40.70            & 0.69          & 0.26          & 0.09          \\
                                          & Direct                                       & 48.50            & 0.61          & 0.39          & 0.10          & 54.51            & 0.58          & 0.44          & 0.15          \\
                                          & HuggingGPT                                   & 48.55            & 0.63          & 0.36          & 0.10          & 55.62            & 0.55          & 0.44          & 0.13          \\
                                          & PLaG                                         & 48.18            & 0.71          & 0.35          & 0.08          & 56.58            & 0.52          & 0.46          & 0.14          \\
                                          & GNN4Plan                                     & 63.66            & 0.47          & 0.53          & 0.23          & 74.44            & 0.33          & 0.69          & 0.48          \\
                                          & GTool                                        & 77.81            & 0.30          & 0.69          & 0.41          & 85.59            & 0.21          & 0.79          & 0.62          \\
                                          & ToolGen                                      & 86.75            & 0.17          & 0.81          & 0.57          & 89.45            & 0.15          & 0.86          & 0.69          \\
                                          & Ours                                         & \textbf{89.93  } & \textbf{0.13} & \textbf{0.84} & \textbf{0.65} & \textbf{92.97  } & \textbf{0.11} & \textbf{0.90} & \textbf{0.76} \\ \hline
\multirow{8}{*}{Mistral-7B}               & BeamSearch                                   & 61.78            & 0.49          & 0.46          & 0.14          & 74.00            & 0.37          & 0.57          & 0.45          \\
                                          & Direct                                       & 55.39            & 0.56          & 0.45          & 0.08          & 65.80            & 0.47          & 0.52          & 0.19          \\
                                          & HuggingGPT                                   & 55.29            & 0.56          & 0.46          & 0.08          & 53.48            & 0.56          & 0.41          & 0.13          \\
                                          & PLaG                                         & 57.08            & 0.54          & 0.48          & 0.09          & 55.62            & 0.55          & 0.40          & 0.13          \\
                                          & GNN4Plan                                     & 58.78            & 0.51          & 0.51          & 0.18          & 72.63            & 0.36          & 0.68          & 0.44          \\
                                          & GTool                                        & 82.85            & 0.23          & 0.72          & 0.49          & 87.53            & 0.19          & 0.80          & 0.62          \\
                                          & ToolGen                                      & 88.04            & 0.14          & 0.82          & 0.62          & 91.18            & 0.14          & 0.88          & 0.72          \\
                                          & Ours                                         & \textbf{89.99  } & \textbf{0.13} & \textbf{0.85} & \textbf{0.65} & \textbf{92.67  } & \textbf{0.11} & \textbf{0.89} & \textbf{0.74} \\ \hline
\end{tabular}}
\end{table*}

\section{Prompts}
\label{app:prompts}

In on-policy distillation, the student and teacher models receive different prompt inputs.
The student prompt contains only the user query and asks the model to directly predict the ordered tool-token sequence.
The teacher prompt additionally receives a reference solution constructed from the ground-truth reasoning subtasks and gold tool sequence.

\paragraph{Student prompt template.}
\begin{verbatim}
System:
Predict the exact ordered tool token sequence needed to complete the user query.
A query may require multiple tools. Output only tool tokens in order, with no
spaces, commas, or other separators between them. Do not explain, do not restate
the query, and do not output any extra words or punctuation.

User:
Query: {query}
\end{verbatim}

\paragraph{Teacher prompt template.}
\begin{verbatim}
System:
Predict the exact ordered tool token sequence needed to complete the user query.
A query may require multiple tools. Output only tool tokens in order, with no
spaces, commas, or other separators between them. Do not explain, do not restate
the query, and do not output any extra words or punctuation. You are given a 
reference solution to help determine the final tool sequence.

User:
Query: {query}
Here is a reference solution:
{reference_solution}
Based on the reference solution above, output the exact correct tool sequence only.
\end{verbatim}

\paragraph{Example from the multimedia dataset.}
For the multimedia sample with ID 25717055, the user query is:
\begin{verbatim}
I have a video file named 'example.mp4', please extract the audio track from it,
then change the pitch to a higher tone according to my instruction 'Increase
pitch'. After that, create a colorful spectrogram image from the modified audio.
\end{verbatim}

The teacher prompt is:
\begin{verbatim}
System:
Predict the exact ordered tool token sequence needed to complete the user query.
A query may require multiple tools. Output only tool tokens in order, with no
spaces, commas, or other separators between them. Do not explain, do not restate
the query, and do not output any extra words or punctuation. You are given a
reference solution to help determine the final tool sequence.

User:
Query: I have a video file named 'example.mp4', please extract the audio track
from it, then change the pitch to a higher tone according to my instruction
'Increase pitch'. After that, create a colorful spectrogram image from the
modified audio.

Here is a reference solution:
Step: 1 Extract the audio track from the input video. Use tool: <Video_to_Audio>
Step: 2 Modify the extracted audio's characteristics according to the user's
instruction. Use tool: <Voice_Changer>
Step: 3 Generate a visual representation of the modified audio track. Use tool:
<Audio_to_Image>
Step: 4 Add color to the visual representation using deep learning techniques.
Use tool: <Image_Colorizer>
Correct tool sequence:
<Video_to_Audio><Voice_Changer><Audio_to_Image><Image_Colorizer>

Based on the reference solution above, output the exact correct tool sequence only.
\end{verbatim}

The expected target sequence for both prompts is:
\begin{verbatim}
<Video_to_Audio><Voice_Changer><Audio_to_Image><Image_Colorizer>
\end{verbatim}

\newpage

\end{document}